\documentclass[journal]{IEEEtran}
\usepackage[pdftex]{graphicx}
\ifCLASSINFOpdf
\else
\fi
\usepackage{url}

\usepackage[ruled,vlined]{algorithm2e}
\usepackage{graphicx}
\usepackage{color}
\usepackage{placeins}
\usepackage{float}
\usepackage{tabularx,colortbl}
\usepackage{multirow}
\usepackage{makecell}
\usepackage{CJK}
\usepackage{amsmath,bm}
\usepackage{cite}
\usepackage{amsthm}
\usepackage{extarrows}
\usepackage{amssymb}
\usepackage[table]{xcolor}

\begin{document}
%

\title{Minimum Noticeable Difference based Adversarial Privacy Preserving Image Generation}

\author{Wen Sun*,
        Jian Jin*,~\IEEEmembership{Member,~IEEE},
       and~Weisi~Lin,~\IEEEmembership{Fellow,~IEEE}

\thanks{Copyright \copyright 20XX IEEE. Personal use of this material is permitted. However, permission to use this material for any other purposes must be obtained from the IEEE by sending an email to pubs-permissions@ieee.org. \emph{*Equal contributions; Corresponding author: Weisi Lin.}}
\thanks{W. Sun, J. Jin and W. Lin are with the School
of Computer Science and  Engineering, Nanyang Technological University, 50 Nanyang Avenue, Singapore, 639798  e-mail: \{sunw0023, jian.jin, wslin\}@ntu.edu.sg.}
\thanks{Manuscript received April 19, 2005; revised August 26, 2021.}}

\markboth{Submit to IEEE Transactions on Circuits and Systems for Video Technology.}{Sun {\it \lowercase{et al.}}: {Minimum Noticeable Difference based Adversarial Image Generation for Privacy Preserving}
}
%



\maketitle

\begin{abstract}

Deep learning models are found to be vulnerable to adversarial examples, as wrong predictions can be caused by small perturbation in input for deep learning models. Most of the existing works of adversarial image generation try to achieve attacks for most models, while few of them make efforts on guaranteeing the perceptual quality of the adversarial examples. High quality adversarial examples matter for many applications, especially for the privacy preserving. In this work, we develop a framework based on the Minimum Noticeable Difference (MND) concept to generate adversarial privacy preserving images that have minimum perceptual difference from the clean ones but are able to attack deep learning models. To achieve this, an adversarial loss is firstly proposed to make the deep learning models attacked by the adversarial images successfully. Then, a perceptual quality-preserving loss is developed by taking the magnitude of perturbation and perturbation-caused structural and gradient changes into account, which aims to preserve high perceptual quality for adversarial image generation. To the best of our knowledge, this is the first work on exploring quality-preserving adversarial image generation based on the MND concept for privacy preserving. To evaluate its performance in terms of perceptual quality, the deep models on image classification and face recognition are tested with the proposed method and several anchor methods in this work. Extensive experimental results demonstrate that the proposed MND framework is capable of generating adversarial images with remarkably improved performance metrics (e.g., PSNR, SSIM, and MOS) than that generated with the anchor methods.
\end{abstract}


%
\IEEEpeerreviewmaketitle

\section{Introduction}
\label{Intro}


\IEEEPARstart{W}{ith} annotated large-scale datasets (\emph{e.g.}, ImageNet\cite{deng2009imagenet}), effective optimizaton algorithms (\emph{e.g.}, Back-propagation\cite{Backpropagation}), as well as the efficient computing resources (\emph{e.g.}, Graphics Processing Unit (GPU)), deep learning has made breakthroughs in many computer vision tasks, such as image classification \cite{deng2009imagenet}, object detection \cite{FasterRCNN}, semantic segmentation \cite{FCN}, image quality assessment~\cite{IQACNN}, and so on. It has been demonstrated that deep learning algorithms even outperform human beings for image classification tasks \cite{He_2015_ICCV}. 


However, the deep learning algorithms are found to make wrong predictions to the adversarial images\cite{PhysicalAttack}, which are usually generated by adding perturbation in the clean image based on different assumptions and conditions. Although the adversarial image and clean one are similar in some degree, the predictions from deep learning system vary significantly. Adversarial images aim to fool the classifier to make wrong decisions, which usually brings threaten to the safety of the practical deep learning applications such as autonomous driving, safety surveillance, and intelligent healthcare. However, it also has positive meanings in privacy preserving applications, such as the application of face photo sharing in online social media. In such a scenario, the adversarial images of user's photos are shared instead of their clean ones, which enables the 
user’s identification information being unrecognizable and preserved without affecting their photo sharing experience. This also requires that the adversarial images should be similar to their clean ones as much as possible. Actually, the average PSNR/SSIM between the adversarial images and their clean ones are only around 30$\sim$35dB/0.78$\sim$0.89 according to our investigation of the anchor methods on ImageNet dataset. Their visual perceptual difference is distinguishable. 
Although some existing adversarial attack methods implicitly enforce that the generated image is not too far away from clean one. For instance, the Fast Gradient Sign Method (FGSM)\cite{AdverExamples} constrains that the generated image is within $\ell_{\infty}$-norm with $\epsilon$ as the perturbation. However, there have been few works on preserving high perceptual quality during adversarial image generation. 

In view of above, we will study how to generate quality-preserving adversarial images in this work. Specifically, we will propose a new objective function to generate the adversarial image that not only can successfully attack the deep learning model, but also preserve the high perceptual quality to make the generated adversarial image with minimum difference for the human visual system. The major contributions of this paper are summarized as follows:
\begin{itemize}
  \item To the best of our knowledge, this is the first work that explores quality-preserving adversarial image generation based on MND concept, where the MND based adversarial image is explicitly defined and formulated. Unlike no exact quality requirements on the adversarial images generated with the other methods, MND based adversarial image is expected to have the best perceptual quality for the human visual system. Hence, the proposed method is more suitable for privacy preserving and relevant applications.  
  
  \item To generate the MND based adversarial images, we firstly propose the adversarial losses for targeted attacks and non-targeted attacks, which guarantees that the adversarial images can successfully attack deep learning models. Meanwhile, we also develop a perceptual quality-preserving loss, which aims to preserve high perceptual quality for adversarial image generation. Hence, factors highly related to perceptual quality, including the magnitude of the perturbation and the perturbation-caused structural and gradient changes, have been considered. 
  
  \item We verify the proposed method together with the anchors on the ImageNet and VGGface2 for image classification and face identification tasks under targeted attacks and non-targeted attacks settings, which include objective quality evaluation (e..g, PSNR, and SSIM) and subjective viewing tests (e.g., MOS) on the adversarial images as well as additional analysis on heatmaps and absolute difference maps. All the extensive experimental results demonstrate that the proposed MND framework is capable of generating adversarial images with remarkably improved performance metrics than that generated with the anchor methods.

\end{itemize}

\section{Related Works}
\subsection{Adversarial attack}
The different types of adversarial attacks can be summarized into two major categories~\cite{AttackSurvey}, including targeted attack and non-targeted attack. The targeted attack is able to trick the classifier to a 
pre-specified target class, while the non-targeted attack just tricks the classifier to make a wrong classification. Besides, there are also black-box attack and white-box attack based on the knowledge of the target model~\cite{AttackSurvey}. In the white-box attack, the details of the model are assumed to be known to the attacker, while in the black-box attack, the attacker has no information on the model.


The problem of the adversarial attack was firstly introduced in~\cite{SzegedyZSBEGF13}, where the existing machine learning methods including the deep learning algorithms were found susceptible to the adversarial attack and a L-BFGS algorithm with a box-constraint was proposed to generate adversarial examples to fool the deep neural networks. After that, the Fast Gradient Sign Method (FGSM)~\cite{AdverExamples} was proposed to get the adversarial examples through a single update using the sign of the gradient. The FGSM was a popular baseline for generating the adversarial examples for deep neural networks, which was further improved to the iterative method called Iterative Fast Gradient Sign Method (I-FGSM) in~\cite{BIM} by updating the image generation in an iterative manner. Then, the I-FGSM was further extended to the momentum version, namely Momentum Iterative Fast Gradient Sign Method (MI-FGSM) in~\cite{MIFGSM}, by integrating the momentum term to stabilize update directions to get rid of poor local maxima during the iterative process. Recently, the Diverse Inputs Iterative
Fast Gradient Sign Method DI$^{2}$-FGSM and Momentum Diverse Inputs Iterative Fast Gradient Sign Method (M-DI$^{2}$-FGSM) were proposed \cite{DI2FGSM} to consider the input diversity strategy, which can improve the transferability of adversarial examples under both black-box and white-box attack settings.

All these adversarial attack methods reviewed above are trying to find a more efficient way or generate more proper adversarial images to successfully fool the deep neural networks. However, none of them takes the perceptual quality of the adversarial images into account, which leads to the obvious perceptual distortion appear in the adversarial images. Such kind of distortion largely reduces our visual experience, which is not friendly for the privacy preserving applications as mentioned in Section \ref{Intro}.

\subsection{Privacy preserving photos sharing}
privacy preserving photos sharing has been a popular topic with the increasing development of online social media. Its applications and relevant technologies have attracted more and more attention in both industrial and academics. Commonly, existing works on privacy preserving mainly include into following categories, i.e., image metadata preservation \cite{PrivacyPhotoSharing}, privacy policy sharing and access authority setting \cite{AccessControl,OSNS}, region of interest (ROI) masking and recovering during image encoding and decoding \cite{VirtualU}, and identification information hiding \cite{PPF}. Identification information hiding aims to preserve the portrait privacy of users or some other significant information while without affecting their photo sharing experience, which is mainly achieved via adversarial attack technologies. On the one hand, significant information (e.g, identification information) is unrecognizable from the shared images. On the other hand,  the image quality should not be undermined. Some related works \cite{FaceAttack} and~\cite{AdvHat} tried to put the virtual eyeglasses on their faces via adversarial attack technologies to prevent them from being recognized. The Generative Adversarial Networks (GAN)~\cite{GAN} were also employed to generate the new adversarial face images. All these attack methods for face recognition are able to satisfy the privacy requirement, however, they cannot be readily used due to modifying the face image attributes, which introduces perturbation that alters the contents of the clean image, and degrades their photo sharing experiences in the online social media. Besides, the clean face image is required to be accessed by the online application, making the risk of leaking the privacy online. The recent work~\cite{PPF} attacked the image at the user end without sacrificing the privacy requirements. However, the image quality has not been taken into account in all these existing works above for adversarial attack for privacy preserving applications. 


\subsection{Perceptual image quality assessment}
\label{SecI-C}
The digital images are vulnerable to a few distortions during the acquisition, compression, transmission, and processing process, leading to the degradation of the image quality in terms of the visual perception. The assessment of the image quality is of great importance for the human visual system (HVS), and the evaluation by the human beings is subjective, time-consuming, and inefficient. Hence, there are great interests from the researchers in developing the quantitative metrics that can be used as the objective measurement for predicting the perceptual image quality. The image quality assessment (IQA) aims to measure the image quality using quantitative metrics that are consistent with human subjective evaluation.

There are quite a lot of traditional image quality assessment metrics that have been developed based on the fixed computational models.
The mean squared error (MSE) and the peak-to-noise ratio (PSNR) are the most simple and popular metrics. The MSE is calculated based on the average of the squared pixel differences between the distorted image and clean image. However, these metrics cannot reflect the perception quality of the human visual system. 
The Structural Similarity Index Measure (SSIM) was proposed in~\cite{IQA} to use the structural information to measure the perceptual quality of the images. The SSIM reflects the similarity between the two images in terms of luminance, contrast, and structure. The Feature Similarity Index Measure (FSIM)~\cite{FSIM} was built upon phase congruency and the image gradient magnitude to characterize the image local quality. 
The Haar Wavelet-Based Perceptual Similarity Index (HaarPSI)~\cite{HaarPSI} was proposed to utilize the coefficients calculated from a Haar wavelet transformation to include local similarities between the two images and the relative importance of image areas.
The Visual Information Fidelity (VIF)~\cite{VIF} is based on natural scene statistics and the notion of image information extracted by the human visual system. Besides, deep learning has been recently applied to learn the image quality assessment metrics~\cite{DISTS, LPIPS}. The Learned Perceptual Image Patch Similarity (LPIPS)~\cite{LPIPS} measures the
Euclidean distance between the representations of neural networks in two images. 
Deep Image Structure and Texture Similarity measure (DISTS)~\cite{DISTS} is proposed based on the pre-trained VGG network to unify the structure and texture similarity. The recent work~\cite{BlindQA} also developed the blind image quality assessment (BIQA) algorithm for cartoon image. In addition, many works \cite{JNDMV, JNDVideo} integrated Just Noticeable Difference (JND) models \cite{JNDMM, jin2022full, JNDEdge} for image/video quality assessment and achieved good performance.


\section{Minimum Noticeable Difference based Adversarial Privacy Preserving Image Generation}


In this section, we firstly give the concepts of Minimum Noticeable Difference (MND) and the MND based adversarial image. Then, the MND based adversarial image generation is formulated as an optimization problem, which is further solved with a detailed optimization algorithm.




\subsection{Definition of the MND based adversarial image}



The {\bf{Minimum Noticeable Difference (MND)}} is defined as \emph{the minimum perceptual difference that the human visual system can distinguish between a clean image and its adversarial one} in this paper. It aims to preserve the quality of adversarial image. Therefore, the {\bf{MND based adversarial image}} can be defined as \emph{the adversarial image that can lead the classifier to generate wrong results while preserving the best perceptual quality for the human visual system.}

There are two properties for MND based adversarial images:
\begin{itemize}
\item \textbf{Property 1 (Adversarial property)}: the classification result of an MND based adversarial image from a pre-trained classifier will be different from the classification result of its clean counterpart. For non-targeted attack, it aims to obtain a different result from the MND based adversarial image compared with its clean one. While for targeted attack, the goal of an MND based adversarial image is to obtain a clearly targeted result (class), specified in advance.  The generated MND image will not be identified by a malicious automatic image recognition system, thus avoiding being used or accessed illegally.
\item \textbf{Property 2 (Perceptual quality-preserving property)}: the MND based adversarial image should look as close as possible to its clean one when they are perceived by humans. Specifically, the difference between MND based adversarial image and its clean one should be minimized in terms of the perception by the HVS. 
\end{itemize}

As aforementioned, most adversarial attack works focus on achieving the adversarial property, while the perceptual quality-preserving property has not been well studied. Hence, the proposed MND based adversarial image is more suitable for privacy preserving and its relevant applications compared with the other methods. As its perceptual quality-preserving property will keep the users' high quality of visual experience. Meanwhile, the adversarial property will preserve the users' important information (e.g., identification information) from being used or accessed illegally. 

\subsection{Formulation of the MND based adversarial image}
In this subsection, we will formulate the MND based adversarial image according to its concept and properties. Before that, let us denote a clean image and its adversarial one as $X$ and $Z$, respectively. We have $X, Z \in \mathbb{R}^{n \times h \times w}$, where $n$, $w$, and $h$ are the channel, width, and height of the image. Assume that a typical CNN based image classification neural network $f(\cdot;\theta)$ (e.g., ResNet \cite{ResNet}, VGG \cite{VGG}, etc.) for $C$ classes is well trained with a commonly used dataset (e.g, ImageNet \cite{deng2009imagenet}). $\theta$ denotes the well trained weights of such a neural network. $f(\cdot;\theta)$ is composed of multiple convolutional layers (containing pooling operation), fully connected layers, and Softmax operation. The predicted result of $X$ and $Z$ are denoted as $y$ and $\hat{y}$ ($y, \hat{y} \in \mathbb{R}^{1 \times C}$), respectively. Then, we have $y = f(X;\theta)$ and $\hat{y} = f(Z;\theta)$. Here, $y$ and $\hat{y}$ are defined as the clean prediction and adversarial prediction, respectively. The associated ground-truth label is denoted as $y^{gt}$, where $y^{gt}\in \mathbb{R}^{1 \times C}$ and $y^{gt}=[{y^{gt}_1,y^{gt}_2,...,y^{gt}_C}]$.

According to the concept of the MND based adversarial image, $Z$ can not only attack an image classifier but also keep the minimum perceptual difference between $Z$ and its clean version $X$. Then, $Z$ can be formulated as follows:
\begin{equation}
\label{eqn:MND}
\arg \min _{Z}\mathcal{L}(Z,X) = \mathcal{L}_{adv}(Z) +  \mathcal{L}_{pqp}(Z,X),
\end{equation}  
where $\mathcal{L}_{adv}$ and $\mathcal{L}_{pqp}$ are two loss terms to be minimized during generating $Z$. $\mathcal{L}_{adv}$ is used to indicate the adversarial property of $Z$, i.e., enforcing that classification result of $Z$ will be far away from that of $X$. $\mathcal{L}_{pqp}$ is used to indicate the perceptual quality-preserving property of $Z$ so that that $Z$ will have the minimum perceptual difference from $X$. More details on definitions of $\mathcal{L}_{adv}(Z)$ and $\mathcal{L}_{pqp}(Z,X)$ will be given in Section \ref{sec:advloss} and Section \ref{sec:perloss}, respectively. 


\subsection{Adversarial loss: $\mathcal{L}_{adv}$}\label{sec:advloss}
The adversarial loss $\mathcal{L}_{adv}$ will be introduced in terms of non-targeted attack and targeted attack, respectively. 

\subsubsection{Non-targeted attack}
As aforementioned, $f(\cdot;\theta)$ is a typical CNN based image classification neural network, which contains convolutional layer (containing pooling operation), and fully connected layer. Let us denote the neural network before Softmax operation and Softmax operation as $f_1(\cdot;\theta)$ and $f_2(\cdot)$, respectively. We have $f(\cdot;\theta) = f_2(f_1(\cdot;\theta))$. Feed $X$ and $Z$ into $f_1(\cdot;\theta)$, we will have the outputs $u$ and $\hat{u}$, where $u = f_1(X;\theta)$ and $\hat{u} = f_1(Z;\theta)$. $u = [u_1,u_2,...,u_C]$, $\hat{u} = [\hat{u}_1,\hat{u}_2,...,\hat{u}_C]$ ($u, \hat{u} \in \mathbb{R}^{1 \times C}$). After applying Softmax operation, we have $y = f_2(u)$ and $\hat{y} = f_2(\hat{u})$, where $y = [y_1,y_2,...,y_C]$ and $\hat{y} = [\hat{y}_1,\hat{y}_2,...,\hat{y}_C]$. For the $i^{th}$ element of $y$ and $\hat{y}$, we have $y_i = f_2(u_i)= \frac{e^{u_i}}{\sum_{k=1}^{C} e^{u_k}}$ and $\hat{y}_i = f_2(\hat{u}_i) = \frac{e^{\hat{u}_i}}{\sum_{k=1}^{C} e^{\hat{u}_k}}$.

For the classification task, Cross-Entropy loss, denoted as $CE(\cdot,\cdot)$ is commonly used to indicate that the predicted result $\hat{y}$ is close to the ground-truth label $y^{gt}$. $CE(\cdot,\cdot)$ is represented as follows,
\begin{equation}\label{eqn:CELoss}
CE(y^{gt},\hat{y}) = -\sum_{i=1}^{C} y^{gt}_i \log(\hat{y}_i).
\end{equation}

In this work, however, instead of minimization the Cross-Entropy loss between $y$ and $y^{gt}$, we need to maximize this loss between $\hat{y}$ and $y$ so that the adversarial result will go away from its associated clean one, leading the generated adversarial image to fool the classifier. However, due to the $\log$ operation, the loss may not be properly backpropagated when the initial prediction $\hat{y}_i$ is close to 1 for the target class\footnote{Using original cross entropy loss will not always lead to successful attack.}. Therefore, in this work, the adversarial loss $\mathcal{L}_{adv}$ for non-targeted attack is represented as follows:
%
\begin{equation}\label{eqn:Nontarget}
\begin{split}
\mathcal{L}_{adv} &= \sum_{i=1}^{C} y_i^{gt} \cdot (\hat{y}_i + \hat{u}_i)\\
&= \langle y^{gt}, (\hat{y} + \hat{u}) \rangle \\
\end{split}
\end{equation}
where $\langle \cdot, \cdot \rangle$ denotes inner product operation. With such an objective function, $\hat{y}_i$ and $\hat{u}_i$ can be penalized directly. 

\subsubsection{Targeted attack} The objective of targeted attack is to generate adversarial image that gives the outcome to be a predefined target result (label), denoted as $y^{t}$. Similarly, we have $y^{t} \in \mathbb{R}^{1 \times R}$ and $y^{t} = [y^{t}_1,y^{t}_2,\ldots,y^{t}_C]$. Besides, target label $y^t$ is different from the clean one $y$. 

In view of this, the adversarial loss for targeted attack is represented as follows:
\begin{equation}\label{eqn:Target}
\begin{split}
\mathcal{L}_{adv} &= -\sum_{i=1}^{C}y_{i}^{t} \cdot (\hat{y_i})^{r}\\
&= - \langle y^t, (\hat{y})^{r} \rangle \\
&= - \langle y^t, (f(Z;\theta))^{r} \rangle
\end{split}
\end{equation}
where $r$ is the power parameter to control the level of regularization. 
The removal of $\log(\cdot)$ operation from standard cross-entropy loss is to avoid the possible zero value for the initial prediction of the target class. The targeted outputs $y^t$ can be predefined using $y = f(X;\theta)$, and in this work, we define it to be the class where the classifier gives the minimum output value for the specific input image $X$, specifically, 
\begin{equation}
\label{255}
 y_{k}^{t}=\left\{
\begin{aligned}
& 1, \text{k=$\arg \min_{m} (y_m)$, m=$1,\ldots,C$}\\
& 0, \text{otherwise}.\\
\end{aligned}
\right.
\end{equation}
It differs from the non-targeted attack in two points: 1) the class under non-targeted attack is uncertain while the class under targeted attack is fixed to be a pre-defined class, and 2) the non-targeted attack is easier to happen than targeted attack.

Therefore, the adversarial loss $\mathcal{L}_{adv}$ can be represented as
\begin{equation}
\label{255}
\mathcal{L}_{adv} = \left\{
\begin{aligned}
&\langle y^{gt}, (f(Z;\theta))^{r} \rangle, \text{if it is non-targeted attack}\\
&- \langle y^t, (f(Z;\theta))^{r} \rangle, \text{if it is targeted attack}.\\
\end{aligned}
\right.
\end{equation}

\subsection{Perceptual quality-preserving loss: $\mathcal{L}_{pqp}$}\label{sec:perloss}
As reviewed in Section \ref{SecI-C}, the perceptual quality of distorted image is highly related to the magnitude of degeneration, namely the perturbation/noise added to the clean image. Besides, the perturbation caused luminance, contrast, structure, and gradient changes are commonly used features, which have been proven their significance for image quality assessment as well. In view of above, three sub-losses are developed to ensure the perceptual quality-preserving property of the adversarial image $Z$ in this work, which is formulated as follows:

\begin{equation}\label{eqn:CELoss}
\begin{split}
\mathcal{L}_{pqp} =& \beta_1 \cdot ||Z-X||_{1} - \beta_2 \cdot SSIM(Z,X)\\ &+\beta_3 \cdot ||g(Z)-g(X)||_{1}.
\end{split}
\end{equation}
The first item $||Z-X||_{1}$ is a deviation loss to penalize the magnitude of perturbation between $Z$ and $X$. The second item $SSIM(Z,X)$ is a structural similarity loss to ensure that $Z$ has the similar structural similarity as $X$ in terms of luminance, contrast and structure. The last item $||g(Z)-g(X)||_{1}$ is a gradient similarity loss to ensure that $Z$ has the similar gradient as $X$. Besides, three hyper parameters $\beta_1$, $\beta_2$ and $\beta_3$ are used to balance such items. The details of three kinds of sub-losses will be introduced below.

\subsubsection{Deviation loss} The deviations, i.e., the magnitude of perturbation between $Z$ and $X$, are usually modelled using the distance between the generated adversarial image $Z$ and the clean one $X$. Here, we generalize the metric of the deviations using the $\ell_1$-norm as below:
\begin{equation}\label{eqn:LpNorm}
||Z-X||_{1}=\sum_{n}\sum_{h}\sum_{w} |Z(n,h,w)-X(n,h,w)|,
\end{equation}
where $||\cdot||_1$ denotes $\ell_1$-norm. $|\cdot|$ denotes absolute value operation. $Z(n,h,w)$ and $X(n,h,w)$ denote the values of pixels locate at $(n,h,w)$ in $Z$ and $X$, respectively. 
Besides, we also study this item with $\ell_2$-norm as below
\begin{equation}
||Z-X||_{2}=\sqrt{\sum_{n}\sum_{h}\sum_{w} (Z(n,h,w)-X(n,h,w))^2}.
\end{equation}
However, according to our experimental results, $\ell_1$-norm is more suitable for penalizing the deviation of $Z$ from $X$ compared with $\ell_2$-norm. More details will be introduced in Section IV.

\subsubsection{Structural similarity loss} As reviewed in Section \ref{SecI-C}, SSIM~\cite{IQA} reflects the similarity between the two images in terms of the luminance, contrast, and structure. Therefore, SSIM is used here to preserve the structural information of the MND image and maintain the high perceptual quality of MND image. SSIM can be written as the below:
\begin{equation}\label{eqn:SSIM}
SSIM(Z,X) = \frac{(2\mu_{z}\mu_{x} + C_{1})(\sigma_{z,x}+C_{2})}{(\mu_{z}^{2} + \mu_{x}^{2} + C_{1})(\sigma_{z}^{2} + \sigma_{x}^{2}+C_{2})},
\end{equation}
where $\mu_{z}, \sigma_{z}$ (\emph{resp.}, $\mu_{x}, \sigma_{x}$) are the mean intensity and standard deviations for the adversarial image $Z$ (\emph{resp.}, $X$) and $\sigma_{z,x}$ is the correlation coefficient that reflects the cosine of the angle between the two images $Z$ and $X$.

Specifically, we have the mean intensity as below:
\begin{equation}
\mu_{z} = \frac{1}{N}\sum_{i} z_{i}, \quad \mu_{x} = \frac{1}{N}\sum_{i} x_{i}
\end{equation}

\begin{equation}
\sigma_{z} = \left(\frac{1}{N-1}\sum_{i} \left(z_{i} - \mu_{z}\right)^2\right)^{1/2},
\end{equation}

\begin{equation}
 \sigma_{x} = \left(\frac{1}{N-1}\sum_{i} \left(x_{i} - \mu_{x}\right)^2\right)^{1/2}
\end{equation}

\begin{equation}
\sigma_{z,x} = \frac{1}{N-1} \sum_{i} \left(x_{i} - \mu_{x}\right) \left(z_{i} - \mu_{z}\right)
\end{equation}

We can see from equation~(\ref{eqn:SSIM}) that the larger the SSIM value, the more similar the MND image $Z$ and clean input $X$ will be. In order to maximize the SSIM value to preserve the perceptual quality of the generated adversarial image, we use the negative value of $SSIM(Z,X)$ for the structural similarity loss in equation~(\ref{eqn:CELoss}).


\subsubsection{Gradient similarity loss}
We also enforce the gradient similarity between the clean image and its associated MND one. To obtain the gradient of $X$ and $Z$, the Sobel operator~\cite{sobel} $g(\cdot)$ is used here. Specifically, for the given input $X$, the Sobel operator is conducted with two steps. Step 1, get the x-direction gradient: 
\begin{equation}
G_x = \big[\begin{smallmatrix}
  +1 & 0 & -1\\
  +2 & 0 & -2\\
  +1 & 0 & -1
\end{smallmatrix}\big]* X.
\end{equation}
Step 2, get the y-direction gradient:
\begin{equation}
G_y = \big[\begin{smallmatrix}
  +1 & +2 & +1\\
  0 & 0 & 0\\
  -1 & -2 & -1
\end{smallmatrix}\big]*X,
\end{equation}
where $*$ is the image convolution operation. The final gradient information from Sobel is as followings,
\begin{equation}
g(X) = \sqrt{G_x^{2} + G_y^{2}}.
\end{equation}
After getting the gradients using Sobel operator for both $X$ and $Z$, the gradient similarity loss is defined as
\begin{equation}
||g(Z)-g(X)||_{1}.
\end{equation}

Similarly,  we also study this item with $\ell_2$-norm as below
\begin{equation}
||g(Z)-g(X)||_{2}.
\end{equation}
Also, we come to the same conclusion that $\ell_1$-norm is more suitable for penalizing the gradient similarity between $Z$ and $X$ compared with $\ell_2$-norm according to our experimental results as to be introduced in Section IV.

It should be noticed that we explicitly include significant features used in image quality assessment into the objective function of the adversarial image generation. Although image quality assessment features, such as deviation, luminance, contrast, structure and gradient have been used for designing various IQA metrics in the existing works to evaluate the image quality, there has been no work to include luminance, contrast, structure and gradient features explicitly into the objective function for adversarial attack work. Experimental results in Section \ref{sec:exp} also demonstrate that such kinds of features are crucial for generating adversarial image with high perceptual quality.

\subsection{Optimization}
The deep learning objective function is usually optimized using the back-propagation algorithm~\cite{Backpropagation}. Therefore, we have to calculate the gradient of the objective function in equation (\ref{eqn:MND}). The gradient of (\ref{eqn:MND}) is calculated as follows:
\begin{equation}\label{eqn:MND_grad}
\begin{split}
\frac{\partial \mathcal{L}(Z,X)}{\partial Z} &=  \frac{\partial \mathcal{L}_{adv}(Z)}{\partial Z} + \frac{\partial \mathcal{L}_{pqp}(Z,X)}{\partial Z}
\end{split}
\end{equation}

From the definition of the $\mathcal{L}_{adv}$ in equation~(\ref{eqn:Nontarget}) (\emph{resp.,} equation~(\ref{eqn:Target})) for non-targeted attack (\emph{resp.,} targeted attack) and $\mathcal{L}_{pqp}$ in equation~(\ref{eqn:CELoss}), we can observe that the partial derivatives of each of the term with respect to $Z$ exists and can be computed during the iterative optimization procedure. Once we have the gradient, the optimization problem can be solved iteratively using the back-propagation algorithm as below:
\begin{equation}\label{eqn:GradUpdate}
Z^{t+1} = Z^{t} - \alpha \frac{\partial \mathcal{L}(Z^{t}, X)}{\partial Z^{t}},
\end{equation}
where $\alpha$ is the step size. 

The algorithm for updating the adversarial image $Z$ can be summarized as below: 
\begin{algorithm}
\SetAlgoLined
\KwResult{The generated optimal MND  image $Z^{\star}$}
 $Z^{0}=X$,\\
 $t = 0$. \\
 \While{ convergence criteria not satisfied}{
  Calculate the gradient of the total objective function (\emph{i.e.} equation~(\ref{eqn:MND})) as in equation~(\ref{eqn:MND_grad}).\\
  Update $Z^{t+1}$ according to equation~(\ref{eqn:GradUpdate})).\\
  $t = t + 1$. \\
 }
 Project the solution $Z^{t}$ into the feasible space.\\
 The final optimum solution $Z^{\star} = Z^{t}$.
 \caption{Optimization algorithm for solving the MND image generation problem in equation~(\ref{eqn:MND})}.
\end{algorithm}

Note that different from the traditional optimization task for optimizing the Convolutional Neural Network, in the adversarial image generation optimization problem, we assume that the Convolutional Neural Network $f(\cdot;\theta)$ has been pre-trained and all the parameters $\theta$ of the Convolutional Neural Network $f(Z;\theta)$ will be frozen. Instead, we are treating the MND image $Z$ as unknown parameters and the gradients will be back-propagated back to $Z$ through the network parameters.





\section{Experiments}\label{sec:exp}

In this section, we make comprehensively comparisons on the image quality of adversarial images, which are generated with the state-of-the-art adversarial attack methods and the proposed one. To this end, we firstly test their performance on the image classification tasks with ImageNet \cite{deng2009imagenet}. Then, the similar experimental settings are carried out and tested on the privacy preserving face identification task with VGGFace2 \cite{VGGFace2}. 

\subsection{Experimental settings}
\subsubsection{Anchors} we compare our proposed MND method with major anchors for adversarial attack. BIM: Basic Iterative Method~\cite{BIM} is the iterative fast gradients sign method (FGSM) method. PGD: Projected Gradient Method~\cite{PGD} initializes the search randomly within the allowed norm ball based on the BIM. MIFGSM: Momentum Iterative Fast Gradient Sign Method~\cite{MIFGSM} integrated the momentum to the BIM model to stabilize the update directions. DI$^{2}$FGSM: Diverse Inputs Iterative Fast Gradient Sign Method~\cite{DI2FGSM} incorporated input diversity strategy with iterative attacks. The torch attacks library~\cite{torchattack}\footnote{\url{https://github.com/Harry24k/adversarial-attacks-pytorch}} is used for implementing the baseline anchors.


\subsubsection{MND algorithms settings} 
To verify the reasonability of our objective function, several different settings of proposed MND algorithm for ablation study are listed below:  
\begin{itemize}
\item \textbf{No norm}, the baseline of proposed MND, the objective function with $\beta_1=\beta_2=\beta_3=0$ in equation (\ref{eqn:CELoss}). Therefore, only the first item is optimized in (\ref{eqn:MND}).

\item \textbf{$\ell_1$-norm}, only the first item of $\ell_1$-norm regularization in equation (\ref{eqn:CELoss}) is kept by setting $\beta_2$ and $\beta_3$ to be 0.

\item \textbf{SSIM}, only the second item of SSIM regularization in equation (\ref{eqn:CELoss}) is kept by setting $\beta_1$ and $\beta_3$ to be 0.

\item \textbf{$\ell_1$-norm + SSIM}, both the first item and the third item are kept by setting $\beta_2$ to be 0 in equation (\ref{eqn:CELoss}).

\item \textbf{MND}, all the three items are kept with $\beta_1, \beta_2$ and $\beta_3$ to be the corresponding values under each setting.

\item \textbf{$\ell_2$-norm}, the $\ell_2$-norm regularization is used in the first item of equation (\ref{eqn:CELoss}) to replace the $\ell_1$-norm. $\beta_2$ and $\beta_3$ are set to be 0.

\item \textbf{$\ell_2$-norm + SSIM}, the $\ell_2$-norm regularization is used in the first item of equation (\ref{eqn:CELoss}) to replace the $\ell_1$-norm. Meanwhile, the first item is abandoned by setting $\beta_1$ to be 0.
\end{itemize} 


\subsubsection{Metrics} the Peak Signal-to-Noise Ratio (PSNR) and Structural Similarity Index (SSIM) are used as the performance metrics to evaluate the image quality of the generated adversarial images from the different methods. Besides, subjective evaluations are also conducted by using Mean Opinion Score (MOS) subjective evaluation. 

\subsection{Image classification}

\subsubsection{Dataset} for the image classification task, we employ the pre-trained deep neural networks with ImageNet dataset. There are 1000 pre-defined classes. The class with the maximum probability from the outputs will be identified as the final prediction.
During the performance evaluation process, we randomly select a total number of 100 images from 10 classes for the test, including tench, goldfish, great white shark, tiger shark, hammerhead, electric ray, stingray, cock, hen and ostrich.


\subsubsection{Backbone Network} 
the VGG19 network \cite{VGG} pre-trained on ImageNet
database for 1000 object classes classification is used as the backbone network. The VGG19 is composed of a
total number of 19 layers, including 16 convolutional layers and 3 fully connected layers. It has achieved promising results for the image classification task. 
For the optimization, the learning rate is fixed to be 0.0001. The maximum number of iterations is set to be 1000. Note that our proposed framework is general and independent of CNN backbone networks and can be applied to any existing pre-trained classifier in a similar way.





\subsubsection{Objective evaluation}
The objective evaluation is conducted under non-targeted and targeted attack settings, respectively. The mean and standard deviation of PSNR and SSIM for all the 100 tested images are shown in TABLE~\ref{table:ImageNet} under non-targeted attack setting. The hyperparameters are set as follows: $\beta_1 = \beta_2 = \beta_3 = 100$. From the results, we can have the following observations.
Firstly, the proposed MND method achieves the best PSNR and SSIM results when compared with anchors, demonstrating the effectiveness of the proposed MND for preserving the quality of generated adversarial images.
Secondly, the deviation loss (\emph{i.e.,} $\ell_1$-norm) is very effective in MND when compared with $\ell_2$-norm and No norm.
Besides, the inclusion of the image quality assessment metric SSIM into the objective function helps to improve the performance of both the PSNR and SSIM scores, demonstrating it is effective to include the structural similarity into the MND. Moreover, the gradient similarity with the Sobel filter is an effective way to improve the image quality further. 
Finally, The final performance from the MND objective based on jointly optimizing the adversarial loss and the perceptual quality-preserving loss gives the best performance.


\begin{table}
\centering
\caption{The performance metric (PSNR and SSIM) in terms of mean $\pm$ standard deviation for the different methods under non-targeted attack for image classification task on ImageNet dataset. }
\label{table:ImageNet}
\begin{tabular}{|c|c|c|}
 \hline
 & \textbf{PSNR} & \textbf{SSIM} \\
 \hline
 \hline
 PGD~\cite{PGD} & 33.21 $\pm$ 0.32 & 0.8481 $\pm$ 0.0709 \\
 \hline
  MIFGSM~\cite{MIFGSM} & 30.92 $\pm$ 0.21 & 0.7790 $\pm$ 0.0951 \\
 \hline
  BIM~\cite{BIM} & 34.85 $\pm$ 0.66 & 0.8961 $\pm$ 0.0465 \\
 \hline
 DI$^{2}$FGSM~\cite{DI2FGSM} & 33.33 $\pm$ 0.30 & 0.8525 $\pm$ 0.0697 \\
 \hline
 \hline
No norm  & 47.56 $\pm$ 2.60 & 0.9809 $\pm$ 0.0138 \\
 \hline
$\ell_2$-norm  & 47.60 $\pm$ 2.56	& 0.9810 $\pm$ 0.0137 \\
\hline
$\ell_1$-norm  & 48.17 $\pm$ 2.12	& 0.9842 $\pm$ 0.0112	 \\
\hline
SSIM & 47.91 $\pm$ 2.44	& 0.9912 $\pm$ 0.0039	 \\
\hline
$\ell_2$-norm + SSIM  & 47.94 $\pm$ 2.42	& 0.9912 $\pm$ 0.0039 \\
\hline
$\ell_1$-norm + SSIM  & 48.40 $\pm$ 2.04	& 0.9922 $\pm$ 0.0037 \\
\hline
MND  & \textbf{48.55} $\pm$ \textbf{1.94}	& \textbf{0.9925} $\pm$ \textbf{0.0036} \\
\hline
\end{tabular}
\end{table}


\begin{table}
\centering
\caption{The performance metrics (PSNR and SSIM) in terms of mean $\pm$ standard deviation for the different methods for image classification task on ImageNet dataset. } 
\label{table:ImageNetTarget}
\begin{tabular}{|c|c|c|}
 \hline
 & \textbf{PSNR} & \textbf{SSIM} \\
 \hline
 \hline
 PGD~\cite{PGD} & 33.76 $\pm$ 0.19 & 0.8567 $\pm$ 0.0714 \\
 \hline
  MIFGSM~\cite{MIFGSM} & 30.76 $\pm$ 0.20 & 0.7692 $\pm$ 0.0997 \\
 \hline
  BIM~\cite{BIM} & 35.68 $\pm$ 0.36 & 0.9036 $\pm$ 0.0489 \\
 \hline
 DI$^{2}$FGSM~\cite{DI2FGSM} & 33.67 $\pm$ 0.22 & 0.8571 $\pm$ 0.0705 \\
 \hline
 \hline
No norm  & 44.19 $\pm$ 2.42 & 0.9535 $\pm$ 0.0261 \\
 \hline
$\ell_2$-norm  & 44.29 $\pm$ 2.33 & 0.9542 $\pm$ 0.0259 \\
\hline
$\ell_1$-norm  & 45.92 $\pm$ 1.87 & 0.9681 $\pm$ 0.0271 \\
\hline
SSIM & 44.36 $\pm$ 2.45	& 0.9601 $\pm$ 0.0218	 \\
\hline
$\ell_2$-norm + SSIM  & 44.48 $\pm$ 2.36	& 0.9609 $\pm$ 0.0219 \\
\hline
$\ell_1$-norm + SSIM  & 46.20 $\pm$ 2.01	& 0.9751 $\pm$ 0.0186 \\
\hline
\textbf{MND}  & \textbf{46.56} $\pm$ \textbf{1.90}	& \textbf{0.9778} $\pm$ \textbf{0.0172} \\
\hline
\end{tabular}
\end{table}



For MND under targeted attack setting, the hyper-parameters are set as: 
$r = 0.0625, \beta_1 = \beta_3 = 2$ and $\beta_2 = 0.1$. The means and standard deviations of PSNR and SSIM  for all the 100 tested images are shown in  TABLE~\ref{table:ImageNetTarget} under the targeted attack setting. From the results, we have similar observations with the non-targeted attack setting. Hence, our proposed MND method can be successfully applied under the targeted attack setting. 
When comparing the results between non-targeted attack and targeted attack, the performance of non-targeted attack is better than targeted attack, demonstrating that non-targeted attack is easier than targeted attack.


\begin{table*}
\centering
\caption{The test results (MOS) on the ImageNet dataset (I1 to I8) under targeted attack setting for MND compared with anchor methods.}
\label{table:MOSResult}
\begin{tabular}{|c|c|c|c|c|c|c|c|c|}
\hline
 & \textbf{VS. BIM}& & \textbf{VS. DI$^{2}$FGSM} && \textbf{VS. MIFGSM} && \textbf{VS. PGD} &  \\
 \hline
 \textbf{Image ID} & \textbf{Mean} & \textbf{Std} & \textbf{Mean} & \textbf{Std}& \textbf{Mean} & \textbf{Std}& \textbf{Mean} & \textbf{Std} \\
 \hline
 \hline
  \textbf{I1} & 0.81 & 1.08 & 1.06 & 0.96 & 1.50 & 1.06& 1.13 & 1.06 \\
 \hline
  \textbf{I2} & 0.75 & 0.94 & 2.13 & 0.86 & 2.50 & 0.63& 2.19 & 0.70 \\
 \hline
  \textbf{I3} & 1.88 & 0.96 & 1.94 & 0.93 & 1.69 & 1.53& 1.56 & 1.06 \\
 \hline
 \textbf{I4} & 2.13 & 0.86 & 2.06 & 0.92 & 2.00 & 1.39 & 2.00 & 0.80 \\
 \hline
 \textbf{I5} & 0.88 & 0.52 & 1.06 & 0.96 & 1.75 & 0.56 & 1.44 & 0.64 \\
 \hline
 \textbf{I6} & 1.75 & 0.68 & 2.38 & 0.64 & 2.31 & 0.63 & 2.06 & 0.74 \\
 \hline
 \textbf{I7} & 2.00 & 0.70 & 2.13 & 0.77 & 2.38 & 0.74 & 1.81 & 1.25 \\
 \hline
 \textbf{I8} & 1.94 & 0.65 & 2.19 & 0.59 & 2.56 & 0.49 & 2.13 & 0.56 \\
 \hline
 \textbf{Average (I1-I8)} & \textbf{1.64}	& - & \textbf{1.87} & - & \textbf{2.09} & - & \textbf{1.79} & - \\
\hline
\end{tabular}
\end{table*}


\begin{table*}
\centering
\caption{The test results (MOS) on the ImageNet dataset (I1 to I8) under non-targeted attack setting for MND compared with anchor methods in terms of mean and standard deviations.}
\label{table:MOSResultNonTargetImageNet}
\begin{tabular}{|c|c|c|c|c|c|c|c|c|c|c|}
\hline
 & \textbf{VS. BIM}& & \textbf{VS. DI$^{2}$FGSM} && \textbf{VS. MIFGSM} && \textbf{VS. PGD} &\\
 \hline
 \textbf{Image ID} & \textbf{Mean} & \textbf{Std} & \textbf{Mean} & \textbf{Std}& \textbf{Mean} & \textbf{Std}& \textbf{Mean} & \textbf{Std}  \\
 \hline
 \hline
 \textbf{I1} & 0.11 & 1.68 & 0.32 & 1.64& 0.16 & 1.70& 0.27 & 1.88 \\
 \hline
 \textbf{I2} & 0.15 & 1.62& 0.06 & 1.87& 0.46 & 1.76& 0.51 & 1.64 \\
 \hline
  \textbf{I3} & 0.43 & 1.51 & 0.22 & 1.74 & 0.70 & 1.62 & 0.26 & 1.74 \\
 \hline
\textbf{I4} & 0.17 & 1.76 & 0.45 & 1.79 & 0.37 & 1.66 & 0.39 & 1.83 \\
 \hline
\textbf{I5}  & 0.10 & 1.69 & 0.10 & 1.67 & 0.36 & 1.47 & 0.23 & 1.60 \\ 
 \hline
\textbf{I6}  & 0.28	& 1.88 & 0.67 & 1.55 & 0.32 & 1.80 & 0.69  & 1.70 \\
\hline
\textbf{I7}  & 0.52	& 1.73 & 0.20 & 1.54 & 0.36 & 1.83 &  0.37 & 1.51  \\
\hline
\textbf{I8} & 0.16	& 1.56 & 0.39 & 1.84& 0.45 & 1.77 & 0.40 & 1.67 \\
\hline
\hline
\textbf{Average (I1-I8)} & \textbf{0.24}	& - & \textbf{0.30} & - & \textbf{0.40} & - & \textbf{0.39} & - \\
\hline
\end{tabular}
\end{table*}

\subsubsection{Details comparison} 
\begin{figure*}
\centering
\includegraphics[width=6.8in]{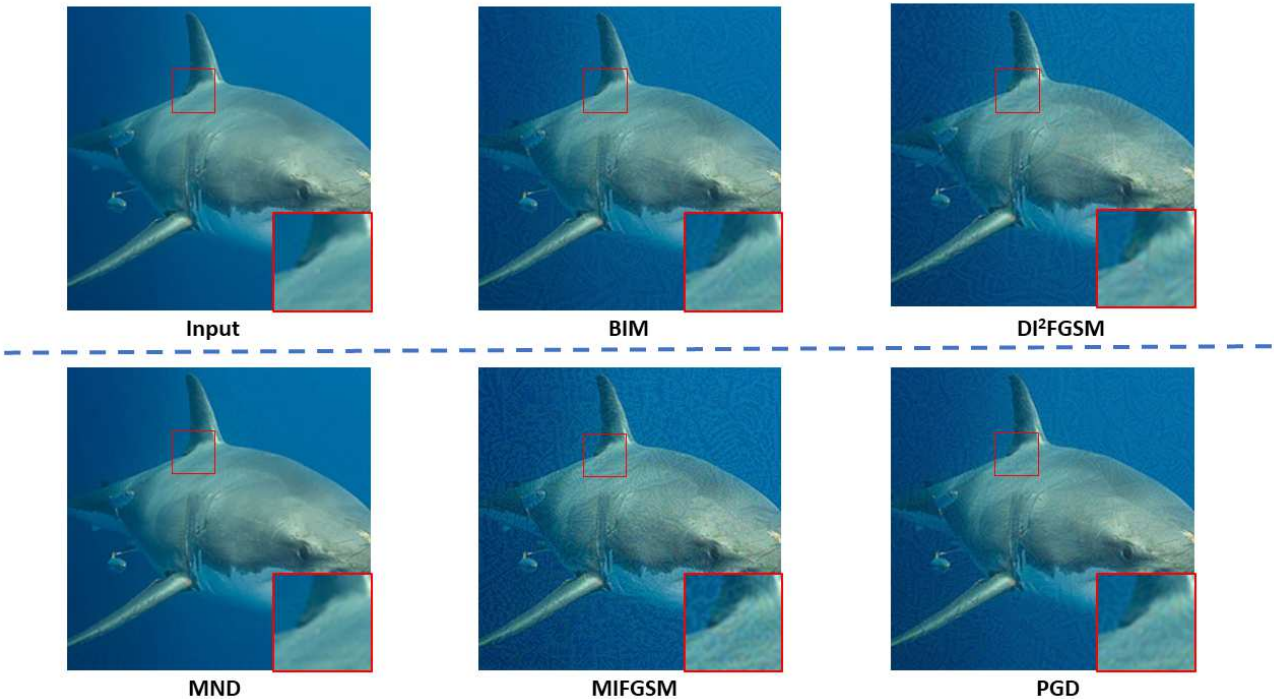}
\caption{
\linespread{2}
Illustration of Minimum Noticeable Difference (MND) image as well as results from anchor methods on the image classification task. The top left image shows the clean input image, and the bottom left shows the MND image, followed by the generated images using the anchors including
BIM, DI$^2$FGSM, MIFGSM and PGD. For each image, both the original one and the enlarged local details (bottom right corner inside the red box) are shown in the figure. }
\label{fig:MNDAdvShark} 
\end{figure*}

One example image from ImageNet dataset under targeted attack with different methods is shown in Fig.~\ref{fig:MNDAdvShark}. The top left image shows the clean input image, followed by the generated images using the anchors including BIM, DI$^{2}$FGSM, PGD, and MIFGSM. According to the visual images in Fig.~\ref{fig:MNDAdvShark}, we can observe that the proposed MND method can generate the adversarial images with minimum noticeable difference with respect to human visual system.


\subsubsection{MOS subjective evaluation}
In order to evaluate the different types of baseline algorithms and the proposed MND algorithm in terms of the perceived image quality objectively from the user perspective, we also conducted the Mean Opinion Score (MOS) subjective evaluation. We follow the guidance of ITU-R BT.500-11 to set up the experimental setting. For each round of subjective evaluation, both the generated images from the proposed MND method and the baseline method will be shown side by side in random order on the screen for marking following the rules defined in TABLE~\ref{table:MOS}. 

\begin{table}
\centering
\caption{Evaluation Standards and Scores for Mean Opinion Score (MOS) Test. }
\label{table:MOS}
\begin{tabular}{|c|c|}
 \hline
 \textbf{Rule} & \textbf{Score} \\
 \hline
 \hline
  Left image is much better & 3 \\
 \hline
  Left image is better & 2 \\
 \hline
 Left image is slightly better & 1 \\
 \hline
Two images are same quality & 0 \\
 \hline
 Right image is slightly better & -1 \\
 \hline
Right image is better	& -2 \\
\hline
Right image is much better	& -3 \\
\hline
\end{tabular}
\end{table}

We randomly selected 8 images (I1 to I8) from ImageNet dataset. The results of the MOS test are reported in TABLE~\ref{table:MOSResult} and TABLE~\ref{table:MOSResultNonTargetImageNet} for targeted attacks and non-targeted attacks, from which we can have the following observations.

Firstly, for the targeted attack on ImageNet dataset, the mean scores of the MOS for MND VS. MIFGSM, MND VS. DI$^{2}$FGSM, and MND VS. PGD and MND VS. BIM are 2.09, 1.87, 1.79 and 1.64, respectively, indicating that the proposed MND is better than these baseline approaches in terms of mean MOS scores. Of all the methods, MND is the best performing method, followed by BIM, PGD, DI$^{2}$FGSM, and MIFGSM. The ranking in terms of the MOS score follows the same order as the results in TABLE~\ref{table:ImageNetTarget} for PSNR scores. 

Secondly, for non-targeted attack on ImageNet dataset, the mean scores of MOS for the MND VS. BIM, MND VS. DI$^{2}$FGSM, MND VS. MIFGSM, and MND VS. PGD are 0.24, 0.30, 0.40 and 0.39, respectively, indicating that the proposed MND is better than these baseline approaches in terms of mean MOS scores. 

These experimental results further demonstrate that our proposed MND framework is good at quality-preserving when compared with the baseline approaches.



\subsection{Face identification}
In this subsection, we will test the proposed methods together with the other anchors on the task of privacy preserving face identification. As mentioned in Section II-B, to achieve privacy preserving face identification, we firstly guarantee that the adversarial images generated by the proposed method and the anchors are able to successfully fool the face identification system. After that, the perceptual quality of the adversarial images generated with the proposed method and the anchors are compared.



\subsubsection{Dataset}
We use the VGGFace2 \cite{VGGFace2} dataset to conduct the experiments. There are a total number of 3.31 million face images from 9131 identities. The dataset was downloaded using Google Image and it contains images with various poses, ages, illuminations, ethnic groups, and professions. A total number of 8631 identities have been used as the training set to train the face identification system. In this work, we focus on the face identification task to identify if the given face image belongs to one of the 8631 identities from VGGFace2 dataset. To demonstrate the effectiveness of the proposed approach, we randomly picked up 100 images from 10 identities out of the total 8631 identities to evaluate the various algorithms.

\subsubsection{Network architecture}
In this work, to demonstrate the application of the proposed MND framework, we employ the popular Multitask Cascaded Convolutional Networks (MTCNN) and a face identification model based on the Inception Resnet (V1) model~\cite{InceptionResnet} pre-trained on VGGFace2 dataset~\cite{VGGFace2}, which have been incorporated in the Facenet-pytorch\footnote{https://github.com/timesler/facenet-pytorch}. Specifically, given one image from VGGFace2, the MTCNN~\cite{MTCNN} face detection model is applied to the image to detect and localize the face. The detected face image is then cropped and normalized to 160$\times$160 pixels since the model has been trained using 160$\times$160 images.

\begin{table}
\centering
\caption{The performance metric (PSNR and SSIM) in terms of mean $\pm$ standard deviation for the different methods under non-targeted attack for face identification task. }
\label{table:FaceNoneTarget}
\begin{tabular}{|c|c|c|}
 \hline
 & \textbf{PSNR} & \textbf{SSIM} \\
 \hline
 \hline
 PGD~\cite{PGD} & 32.94 $\pm$ 0.22 & 0.8671 $\pm$ 0.0133 \\
 \hline
  MIFGSM~\cite{MIFGSM} & 31.96 $\pm$ 0.12 & 0.8451 $\pm$ 0.0165 \\
 \hline
  BIM~\cite{BIM} & 34.31 $\pm$ 0.48 & 0.9054 $\pm$ 0.0080 \\
 \hline
 DI$^{2}$FGSM~\cite{DI2FGSM} & 32.94 $\pm$ 0.21 & 0.8764 $\pm$ 0.0127 \\
 \hline
 \hline
No norm  & 34.31 $\pm$ 1.99 & 0.8329 $\pm$ 0.0405 \\
 \hline
$\ell_2$-norm  & 34.50 $\pm$ 1.91	& 0.8384 $\pm$ 0.0379 \\
\hline
$\ell_1$-norm  & 36.13 $\pm$ 1.64	& 0.8842 $\pm$ 0.0238	 \\
\hline
SSIM & 37.34 $\pm$ 1.73	& 0.9601 $\pm$ 0.0052	 \\
\hline
$\ell_2$-norm + SSIM  & 37.42 $\pm$ 1.70	& 0.9604 $\pm$ 0.0051 \\
\hline
$\ell_1$-norm + SSIM  & 38.20 $\pm$ 1.60	& 0.9638 $\pm$ 0.0048 \\
\hline
MND  & \textbf{38.39} $\pm$ \textbf{1.58}	& \textbf{0.9651} $\pm$ \textbf{0.0047} \\
\hline
\end{tabular}
\end{table}

\begin{table}
\centering
\caption{The performance metric (PSNR and SSIM) in terms of mean $\pm$ standard deviation for the different methods for the face identification task.}
\label{table:FaceID}
\begin{tabular}{|c|c|c|}
 \hline
 & \textbf{PSNR} & \textbf{SSIM} \\
 \hline
 \hline
 PGD~\cite{PGD} & 32.63 $\pm$ 0.26 & 0.8613 $\pm$ 0.0132\\
 \hline
  MIFGSM~\cite{MIFGSM} & 31.78 $\pm$ 0.10 & 0.8421 $\pm$ 0.0169 \\
 \hline
  BIM~\cite{BIM} & 33.38 $\pm$ 0.48 & 0.8872 $\pm$ 0.0094 \\
 \hline
 DI$^{2}$FGSM~\cite{DI2FGSM} & 32.58 $\pm$ 0.01 & 0.8709 $\pm$ 0.0126 \\
 \hline
 \hline
No norm  & 32.90 $\pm$ 1.45 & 0.7893 $\pm$ 0.0356 \\
 \hline
$\ell_2$-norm  & 33.10 $\pm$ 1.37	& 0.7965 $\pm$ 0.0323 \\
\hline
$\ell_1$-norm  & 34.93 $\pm$ 1.27	& 0.8545 $\pm$ 0.0262	 \\
\hline
SSIM & 33.92 $\pm$ 1.33	& 0.8578 $\pm$ 0.0287	 \\
\hline
$\ell_2$-norm + SSIM  & 34.15 $\pm$ 1.30	& 0.8628 $\pm$ 0.0299 \\
\hline
$\ell_1$-norm + SSIM  & 36.15 $\pm$ 1.42	& 0.9048 $\pm$ 0.0336 \\
\hline
\textbf{MND}  & \textbf{36.84} $\pm$ \textbf{1.48}	& \textbf{0.9176} $\pm$ \textbf{0.0330} \\
\hline
\end{tabular}
\end{table}



\begin{figure*}
\centering
\includegraphics[width=6.8in]{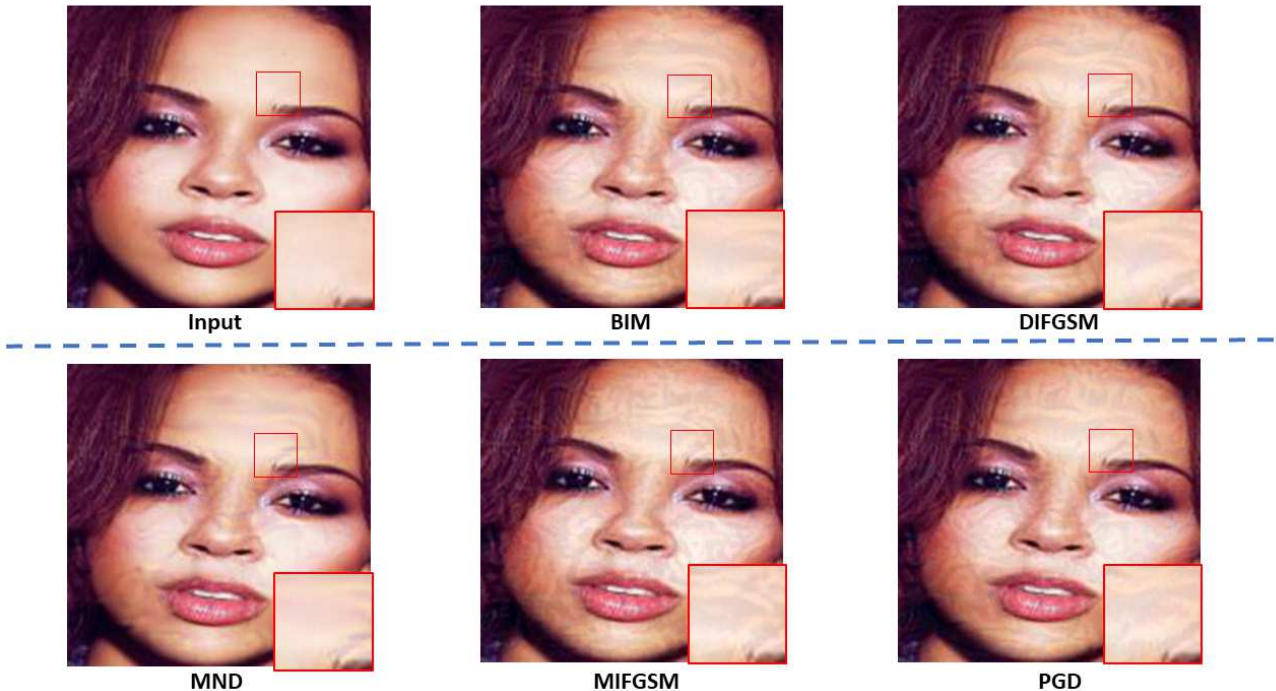}
\caption{
\linespread{2}
Illustration of Minimum Noticeable Difference (MND) image compared with anchor methods on the face recognition task. The top left image shows the clean input image, and the bottom left shows the MND image, followed by the generated images using the anchors including
BIM, DI$^2$FGSM, MIFGSM, and PGD. For each image, both the original one and the enlarged local details (bottom right corner inside the red box) are shown in the figure.}
\label{fig:face_mnd}
\end{figure*}

\begin{table*}
\centering
\caption{The test results (MOS) on VGGFace2 dataset (I9 to I16) under targeted attack setting for MND compared with anchor methods.}
\label{table:MOSResultVGGFace}
\begin{tabular}{|c|c|c|c|c|c|c|c|c|}
\hline
 & \textbf{VS. BIM}& & \textbf{VS. DI$^{2}$FGSM} && \textbf{VS. MIFGSM} && \textbf{VS. PGD} &  \\
 \hline
\textbf{Image ID} & \textbf{Mean} & \textbf{Std} & \textbf{Mean} & \textbf{Std}& \textbf{Mean} & \textbf{Std}& \textbf{Mean} & \textbf{Std} \\
\hline
\hline
 \textbf{I9} & 1.69 & 0.59& 1.31 & 0.82& 1.81 & 0.74& 1.50 & 0.64 \\
 \hline
 \textbf{I10} & 1.13 & 0.64& 1.31 & 0.62& 1.44 & 0.74& 1.31 & 0.90 \\
 \hline
  \textbf{I11} & 0.5 & 0.64& 0.44 & 0.83& 1.06 & 0.88& 1.06 & 0.59\\
 \hline
\textbf{I12} & 0.94 & 0.46& 0.69 & 0.98& 1.44 & 0.74& 1.13 & 0.92\\
 \hline
\textbf{I13}  & 0.19 & 0.83& 0.81 & 0.86& 1.19 & 0.77& 0.19 & 0.83\\ 
 \hline
\textbf{I14}  & 0.63	& 0.74 & 0.56 & 1.06& 1.56 & 0.74& 0.88 & 0.83\\
\hline
\textbf{I15}  & 1.06	& 0.80& 1.25 & 0.70& 1.38 & 1.06& 1.38 & 0.91	 \\
\hline
\textbf{I16} & 0.56	& 0.92& 1.19 & 0.68& 0.69 & 0.98& 0.63 & 0.74 	 \\
\hline
\textbf{Average (I9-I16)} & \textbf{0.84}	& - & \textbf{0.95} & - & \textbf{1.32} & - & \textbf{1.01} & - \\
\hline
\end{tabular}
\end{table*}


\begin{table*}
\centering
\caption{The test results (MOS) on the VGGFace2 dataset (I9 to I16) under non-targeted attack setting for MND compared with anchor methods in terms of mean and standard deviations.}
\label{table:MOSResultNonTargetVGGFace}
\begin{tabular}{|c|c|c|c|c|c|c|c|c|c|c|}
\hline
 & \textbf{VS. BIM}& & \textbf{VS. DI$^{2}$FGSM} && \textbf{VS. MIFGSM} && \textbf{VS. PGD}  &  \\
\hline
 \textbf{Image ID} & \textbf{Mean} & \textbf{Std} & \textbf{Mean} & \textbf{Std}& \textbf{Mean} & \textbf{Std}& \textbf{Mean} & \textbf{Std}  \\
 \hline
 \hline
  \textbf{I9} & 0.46 & 1.58 & 0.19 & 0.67 & 0.26 & 1.53& 0.42 & 1.79  \\
 \hline
  \textbf{I10} & 0.13 & 1.45 & 0.12 & 1.67 & 0.09 & 1.69& 0.19 & 1.78 \\
 \hline
  \textbf{I11} & 0.17 & 1.68 & 0.13 & 1.62 & 0.40 & 1.61& 0.18 & 1.62 \\
 \hline
 \textbf{I12} & 0.25 & 1.57 & 0.22 & 1.83 & 0.11 & 1.66 & 0.22 & 1.55  \\
 \hline
 \textbf{I13} & 0.40 & 1.56 & 0.39 & 1.32 & 0.07 & 1.48 & 0.21 & 1.49  \\
 \hline
 \textbf{I14} & 0.29 & 1.28 & 0.12 & 1.63 & 0.05 & 1.49 & 0.36 & 1.61  \\
 \hline
 \textbf{I15} & -0.02 & 1.65 & 0.32 & 1.72 & 0.33 & 1.68 & 0.19 & 1.55  \\
 \hline
 \textbf{I16} & 0.02 & 1.57 & 0.03 & 1.82 & 0.35 & 1.65 & 0.33 & 1.53  \\
 \hline
 \textbf{Average (I9-I16)} & \textbf{0.21}	& - & \textbf{0.19} & - & \textbf{0.21} & - & \textbf{0.26} & - \\
 \hline
\end{tabular}
\end{table*}

\subsubsection{Objective evaluation}
Similarly, the objective evaluation is conducted under non-targeted and targeted attack settings, respectively. The mean and standard deviation of PSNR and SSIM
for all the 100 tested images are shown in TABLE~\ref{table:FaceNoneTarget} under the non-targeted attack setting. The hyper-parameters are set as: 
$\beta_1 = \beta_2 = \beta_3 = 100$. We have the following observations. Firstly, the proposed MND method achieves the best PSNR (\emph{i.e.,} 38.39) and SSIM (\emph{i.e.,} 0.9651) when compared with all anchor methods including PGD, MIFGSM, BIM and DI$^{2}$FGSM, demonstrating the effectiveness of the proposed MND for preserving the quality of generated adversarial images. Therefore, user privacy with enhanced perception by the human visual system can be preserved accordingly. Secondly, the proposed MND method also surpasses all the special cases of the MND, including No norm, $\ell_2$-norm, $\ell_1$-norm, SSIM, $\ell_2$-norm + SSIM, and $\ell_1$-norm + SSIM. This demonstrates that it is important to incorporate the deviation loss, the structural similarity loss and the gradient similarity loss when generating the adversarial examples under non-targeted attack setting. 

For MND under targeted attack setting for face identification task, the hyper-parameters are set as: 
$r = 0.0625, \beta_1 = \beta_3 = 1$ and $\beta_2 = 0.1$. We report the means and standard deviations of PSNR and  SSIM  for  all  the  100  tested  images  in  TABLE~\ref{table:FaceID} under targeted attack setting. We have the following observations. 
%
%
%
Firstly, the proposed MND method achieves the best PSNR (\emph{i.e.,} 36.84) and SSIM (\emph{i.e.,} 0.9176) compared with baseline methods including PGD, MIFGSM, BIM and DI$^{2}$FGSM, demonstrating the effectiveness of the proposed MND framework for preserving the quality of generated adversarial images. Therefore, user privacy with enhanced perception by human visual system can be preserved accordingly. Secondly, the experimental results demonstrate the effectiveness of the deviation loss, the structural similarity loss and the gradient similarity loss when generating the adversarial examples under targeted attack setting for face recognition task.
\subsubsection{Details comparison} Similarly, one example image from VGGFace2 dataset under targeted attack is shown in Fig. \ref{fig:face_mnd}. The adversarial image generated with the proposed method has the minimum noticeable difference with the original one. In other words, compared with the anchors, the proposed method is able to generate adversarial images with higher perceptual quality.

\subsubsection{MOS subjective evaluation}
We randomly selected 8 images (I9 to I16) from VGGFace2 dataset. The results of the MOS test are reported in TABLE~\ref{table:MOSResultVGGFace} and TABLE~\ref{table:MOSResultNonTargetVGGFace} for targeted attacks and non-targeted attacks, from which we can have the following observations.

Firstly, for targeted attack on VGGFace2 dataset for privacy preserving application, the mean scores of the MND VS. MIFGSM,  MND VS. PGD, MND VS. DI$^{2}$FGSM and MND VS. BIM are 1.32, 1.01, 0.95 and 1.84, respectively, indicating that the proposed MND is better than these approaches in terms of mean MOS scores. Of all the methods, MND is the best performing method, followed by BIM, DI$^{2}$FGSM, PGD and MIFGSM. The ranking in terms of the MOS score follows the same order with the results in TABLE~\ref{table:FaceID} for SSIM scores. 

Secondly, for targeted attack on VGGFace2 dataset, the mean scores of the MND VS. BIM, MND VS. DI$^{2}$FGSM, MND VS. MIFGSM,  MND VS. PGD and MND VS. RFGSM are 0.21, 0.19, 0.21, 0.26 and 0.29, respectively, indicating that the proposed MND is better than these baseline approaches in terms of mean MOS scores. Therefore, for all the methods, MND is the best performing method.

These experimental results further demonstrate that our proposed MND framework is good at quality-preserving when compared with the baseline approaches.

\subsection{Additional analysis}

\subsubsection{Heatmap analysis}
\begin{figure*}
\centering
\includegraphics[width=6.8in]{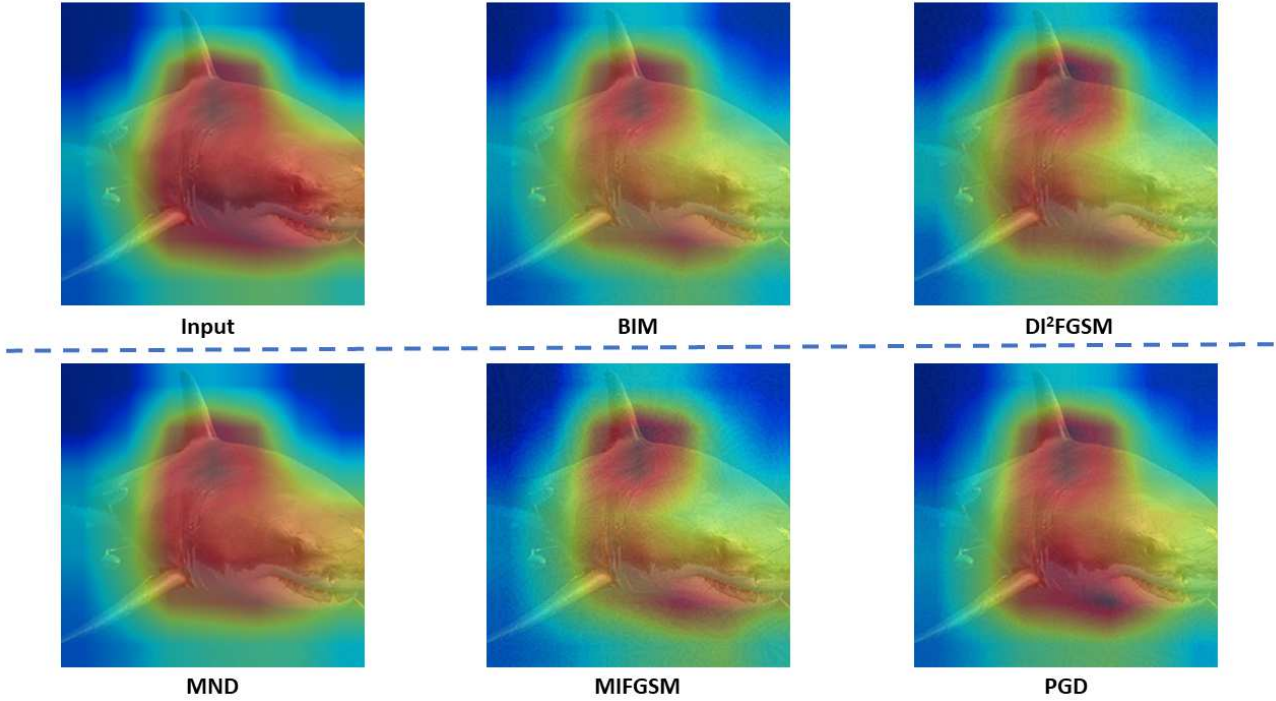}
\caption{
\linespread{2}
Illustration of the heatmaps for Minimum Noticeable Difference (MND) image as well as results from anchor methods under targeted attack setting. The top left image shows the heatmap for the input image, and the bottom left shows the heatmap for the MND image, followed by the generated heatmaps using the anchors including
BIM, DI$^2$FGSM, MIFGSM and PGD.}
\label{fig:imagenet_mnd_heatmap} 
\end{figure*}

To further analyze the adversarial images generated by different methods, the attention of the classifier is represented with heatmap via Gradient-weighted Class Activation Mapping (Grad-CAM) \cite{selvaraju2017grad} technique as shown in Fig.~\ref{fig:imagenet_mnd_heatmap}. 
It can be observed that the attention of the classifier focuses on the target region of the original input image at first. Then, after different kinds of attacks, the attention of the classifier changes slightly in different adversarial images. This demonstrates that different adversarial attack methods change the original image by adding different perturbations. The added different perturbations lead to the attention of the classifier shifting differently. 


\begin{figure*}
\centering
\includegraphics[width=6.8in]{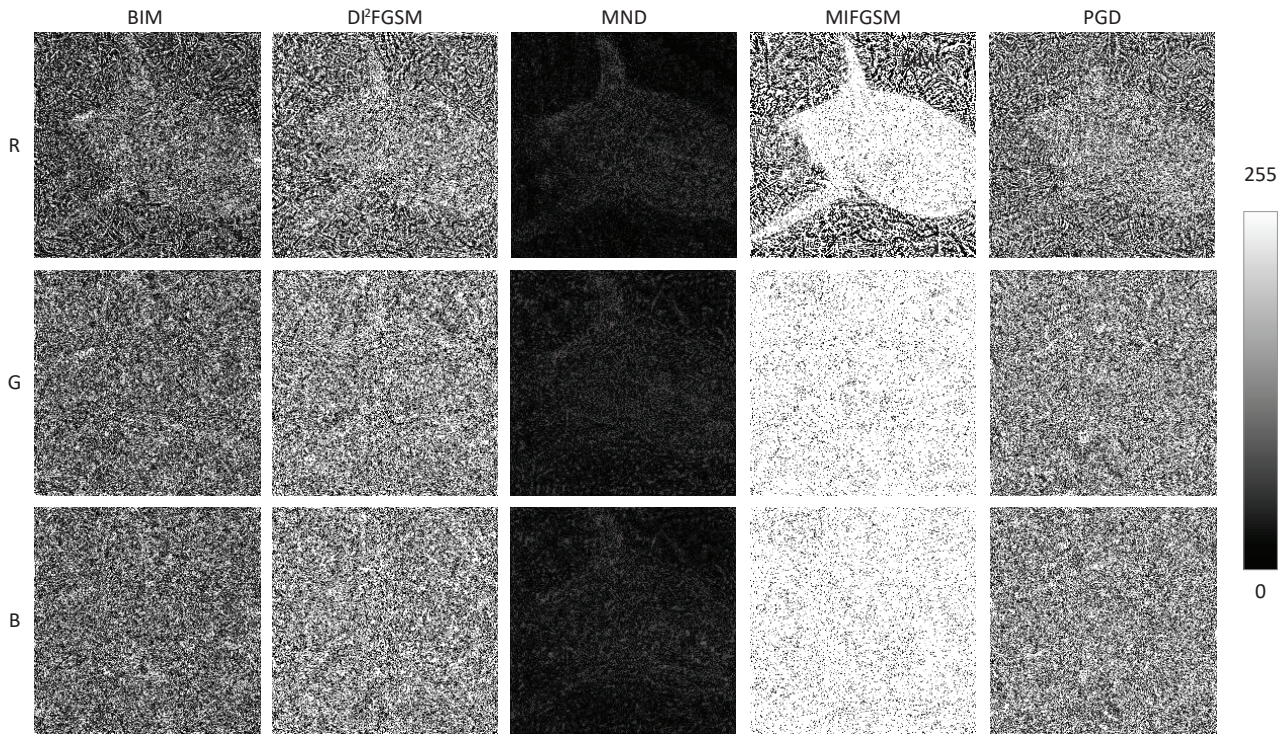}
\caption{
\linespread{2}
Illustration of absolute difference maps on each color channel (e.g., red, green, and blue) between Minimum Noticeable Difference (MND) image (\emph{resp.} anchor methods) and original image under targeted attack setting. For a fair comparison, all the absolute difference maps are normalized to 0$\sim$255, where 0 and 255 are represented with black and white color. The proposed MND image is able to achieve successful attack by adding the minimum perturbation magnitude in three color channels, as compared with the anchor methods.}
\label{fig:imagenet_mnd_diff} 
\end{figure*}


\subsubsection{Absolute difference map analysis}
To further analyze the deviations of the adversarial attack images (generated with different adversarial attack methods) from the original one, we also obtain the absolute difference (i.e., $l_1$-norm) between the adversarial images and their original image of each color channel (e.g., red, blue, and green), as shown in Fig.~\ref{fig:imagenet_mnd_diff}. For a fair comparison, all the absolute difference maps are normalized to 0$\sim$255, where 0 and 255 are represented with black and white color. Based on the observation of Fig.~\ref{fig:imagenet_mnd_diff}, we find that the absolute difference maps of the proposed MND are the darkest. It demonstrates that the proposed MND image is able to achieve a successful attack by adding the minimum perturbation magnitude in three color channels, as compared with the anchor methods. 

Besides, we also calculate the ratio of deviation pixel for all the adversarial images generated with the proposed MND and the other anchors. The ratio is defined as $\frac{\# \text{non-zero}}{N}$, where $N$ is the total number of pixels in the absolute difference map and $\# \text{non-zero}$ is the total number of non-zero pixels in the absolute difference map. The ratio of deviation pixel for the image in Fig.~\ref{fig:imagenet_mnd_diff} of BIM, MIFGSM, DI$^{2}$FGSM, PGD and MND are 0.8240, 0.9491, 0.8240, 0.9122 and 0.6345, respectively. This demonstrates that the proposed MND can achieve a successful attack by changing fewer number of pixels compared with the other anchors. In other words, more information from the original image can be preserved, reducing the probability of image quality degradation due to pixel deviation.


\section{Conclusions}
In this work, we have proposed a framework based on Minimum Noticeable Difference (MND) to generate adversarial privacy preserving images that can attack deep learning based systems but yet to be as close as possible to the clean input so that it has the minimum perceptual difference to the human visual system. Specifically, a new loss function incorporating the adversarial loss and perceptual quality-preserving loss (made up of three sub-losses, i.e., the deviation loss, the structural similarity loss, and the gradient similarity loss) has been proposed. The proposed MND together with the anchor methods has been widely verified on both the ImageNet and the VGGFace2 datasets for image classification and face identification tasks, under both the non-targeted attack and targeted attack settings. Experimental results demonstrate that the proposed methods are able to generate the adversarial images with the highest perceptual quality in terms of PSNR, SSIM, and MOS, while successfully attacking the deep learning models. Therefore, the proposed method is more suitable for privacy preserving and relevant applications.  

\end{document}